\documentclass[aps,prl,twocolumn,showpacs,floatfix,superscriptaddress]{revtex4-1}

\usepackage{times,mathrsfs}
\usepackage{graphicx,color}
\usepackage{amsmath,amssymb}
\usepackage[colorlinks,citecolor=red]{hyperref}

\begin{document}

\newcommand\hatH{{\hat{H}}}
\newcommand\bfJ{{\bf{J}}}
\newcommand\bfK{{\bf{K}}}
\newcommand\bfX{{\bf{X}}}
\newcommand\bfY{{\bf{Y}}}
\newcommand\bfU{{\bf{U}}}
\newcommand\bfV{{\bf{V}}}
\newcommand\bfQ{{\bf{Q}}}

\newcommand\calZ{{\mathcal{Z}}}
\newcommand\calF{{\mathcal{F}}}
\newcommand\calG{{\mathcal{G}}}
\newcommand\avg[1]{\mathinner{\langle{\textstyle#1}\rangle}}
\newcommand\set[1]{\mathinner{\{\textstyle#1}\}}
\newcommand\dblbra[1]{\langle\!\langle #1\rangle\!\rangle}

\newcommand\obs{{\mathrm {obs}}}
\newcommand\eps{{\epsilon}}
\newcommand\KL{{\mathrm {KL}}}
\newcommand\model{{\mathrm {mod}}}

\makeatletter
\newcommand{\printfnsymbol}[1]{%
	\textsuperscript{\@fnsymbol{#1}}%
}
\makeatother

\newcommand\red{\color{red}}
\newcommand\blue{\color{blue}}
\newcommand\gray{\color{gray}}
\definecolor{gray}{gray}{0.8}
\newcommand\rev{\color{red}}

\title{Inverse Ising inference from high-temperature re-weighting of observations}

\author{Junghyo Jo}
\email[Corresponding author: ]{jojunghyo@snu.ac.kr}
\affiliation{Department of Statistics, Keimyung University, Daegu 42601, Korea}
\affiliation{School of Computational Sciences, Korea Institute for Advanced Study, Seoul 02455, Korea}
\affiliation{Department of Physics Education, Seoul National University, Seoul 08826, Korea}

\author{Danh-Tai Hoang}
\affiliation{Laboratory of Biological Modeling, National Institute of Diabetes and Digestive and Kidney Diseases, National Institutes of Health, Bethesda, Maryland 20892, USA}
\affiliation{Department of Natural Sciences, Quang Binh University, Dong Hoi, Quang Binh 510000, Vietnam}

\author{Vipul Periwal}
\email[Corresponding author: ]{vipulp@mail.nih.gov}
\affiliation{Laboratory of Biological Modeling, National Institute of Diabetes and Digestive and Kidney Diseases, National Institutes of Health, Bethesda, Maryland 20892, USA}

\date{\today}

\begin{abstract}
Maximum Likelihood Estimation (MLE) is the bread and butter of system inference for stochastic systems. In some generality, MLE will converge to the correct model in the infinite data limit. In the context of physical approaches to system inference, such as Boltzmann machines, MLE requires the arduous computation of partition functions summing over all configurations, both observed and unobserved. 
We present here a conceptually and computationally transparent data-driven approach to system inference that is based on the simple question: How should the Boltzmann weights of observed configurations be modified to make the probability distribution of observed configurations close to a flat distribution? This algorithm gives accurate inference by using only observed configurations for systems with a large number of degrees of freedom where other approaches are intractable.
\end{abstract}


\maketitle

{\it Introduction.} 
Inferring underlying models from observed configurations is a general task for machine learning.
Maximum Likelihood Estimation (MLE) is a mathematically rigorous approach to parameter estimation for stochastic systems. If a system is observed in configuration $\sigma$ with frequency $n_\sigma$ in a set of $N$ observations, then MLE posits that an estimate of the true probabilities $\{p_\sigma\}$ is 
\begin{equation}
\label{eq:MLE}
\{p^*_\sigma\} = \arg\max_{\{p_\sigma\}} \prod_\sigma p^{n_\sigma}_\sigma .
\end{equation}
Taking the constraint $\sum_\sigma p_\sigma = 1$ into account, one obtains the intuitive result
\begin{equation}
\label{eq:MLEsolve}
p^*_\sigma = {n_\sigma \over {N}} \equiv f_\sigma.
\end{equation}

Ising and Potts models, known as Markov random fields or undirected graphical models in the machine learning and statistical inference fields, are important classes of physical models to represent $p_\sigma$ of observed configurations $\sigma$.
In particular, the models have been adopted to explain neural activities~\cite{schneidman2006, cocco2009, watanabe2013}, gene expression levels~\cite{lezon2006}, protein structures~\cite{weigt2009, cocco2018}, gene recombinations~\cite{mora2010}, bird interactions~\cite{bialek2012}, financial markets~\cite{bury2013, borysov2015}, and human interactions~\cite{eagle2009}.
Searching in the space of graph structures encoding interactions between variables is an NP-hard problem~\cite{chickering1996}.

As a concrete example, for a dataset comprised of $N$ observed configurations of $M$ binary variables $\sigma_i=\pm1,$ the binary variables are associated with Ising spins and the probability of observing a specific configuration $\sigma = (\sigma_1, \sigma_2, \cdots, \sigma_M)$ is assumed to be the normalized Boltzmann weight:
\begin{equation}
\label{eq:Boltzmann}
p_\sigma(w) = {\exp(w^IO_I(\sigma)) \over Z(w)} \ \hbox{\rm with}\ Z(w) \equiv \sum_\sigma  \exp(w^IO_I(\sigma)),
\end{equation}
using the Einstein summation convention between repeated raised and lowered indices, where $\{O_I\}$ is a set of operators appropriate for the problem of interest, for example the set of products $\{\sigma_i\sigma_j, i<j\}.$ The inference problem is to determine the parameters $w^I$ from the data.  Applying MLE estimation, we wish to find $p^*_\sigma$ that maximizes the likelihood ${\cal L}\equiv \prod_\sigma p^{f_\sigma}_\sigma = \prod_{\hat{\sigma}} p_{\hat{\sigma}}^{f_{\hat{\sigma}}}$ as the frequency $f_\sigma$ of unobserved configurations is 0. 
Note that $\hat{\sigma}$ represents observed configurations in $\{ \sigma\}$.
Taking the logarithm of ${\cal L},$ we find 
\begin{equation}
\label{eq:BMgradient}
{\partial{\ln\cal L}\over {\partial w^{I}}} = \langle O_I\rangle_f - \langle O_I\rangle_p.
\end{equation}
Here, for any observable $O_I$ defined on the set of all configurations, $\langle O_I \rangle_f \equiv \sum_{\hat{\sigma}} f_{\hat{\sigma}} O_I(\hat{\sigma})$ is summed over the set of observed configurations, and the model prediction $\langle O_I \rangle_p \equiv \sum_\sigma p_\sigma(w) O_I(\sigma)$ is summed over all configurations. The coupling dependence is entirely in $\langle O_I\rangle_p.$ Gradient ascent using Eq.~(\ref{eq:BMgradient}) to find $w$ has the usual issues with local maxima but the most computationally intensive part is the evaluation of $\langle O_I\rangle_p$ for every step. No matter the size of the available data, $N,$ this computation is a sum with $2^M$ terms.

The computational intractability of the partition function is well-known to physicists~\cite{welsh1993}. 
Due to the centrality of this inverse problem, many approximate solutions have been developed~\cite{nguyen2017}, including machine learning with variational autoregressive networks~\cite{wu2019}.
First and second moments of the data are sufficient statistics to solve the problem, but attempts to use this information alone give inaccurate results for large numbers of spins, suggesting the use of higher moments to improve inference. The adaptive cluster expansion uses heuristics to truncate likelihood computations~\cite{cocco2011, nguyen2012}, and the probabilistic flow method uses relaxation dynamics to aim for pre-specified analytically tractable target distributions, extracting information about the true distribution from reversed dynamics~\cite{sohl2011}. However, both approaches are computationally expensive and thus not applicable for large systems.

It has become clear over a decade of work that methods based on logistic regression perform much better for strongly coupled interactions than mean-field approaches. Many of these approaches use regularized pseudo-likelihood estimators assuming local interaction graphs with restricted connectivity~\cite{ravikumar2010, aurell2012}.
The pseudo-likelihood attempts to circumvent the difficulty of computing the exact partition function. The initial work of Ravikumar et al.~\cite{ravikumar2010} provides incorrect inference for large couplings but
recent improvements~\cite{lokhov2018} have found extensions to this regime as well and achieved very good performance on graphs with limited average degrees, necessary for the locality assumption underlying this regularized approach. In particular, their estimation procedure sets a threshold for small couplings, infers an interaction graph and then learns the values of couplings set to zero by the regularization but only for  the graph structure already inferred. 
Decelle and Ricci-Tersenghi~\cite{decelle2014} showed that the local character of the pseudo-likelihood leads to inaccuracies for the interaction inferred between two spins if their local neighborhoods lead to very different estimates. They avoid this problem by using decimation and obtain excellent results for graphs with bounded degree distributions.

Of course, when we are faced with an inference problem, we have no way of knowing if the couplings are large or small, or if the interaction graphs of the spins have dense or sparse connectivity or have heavy-tailed degree distributions.
Our aim in this Letter is to rethink model inference for such problems to simplify the calculation of $Z(w)$ using elementary considerations. 
We show that this entirely data-driven algorithm is computationally very fast, and accurate even in the hard inference regime of strong coupling with small number of samples. 
Complete source code with documentation is available on GitHub~\cite{Jo2019}.


{\it Theory.} 
The key idea is to trivialize observed configurations by re-weighting their frequencies to make every configuration equally likely. Then, the partition function becomes trivially computable as $\tilde{Z} \approx 2^M$.
Suppose we re-weight $f_\sigma$ by multiplying $q_\sigma^{\epsilon -1}$ of an arbitrary distribution $q_\sigma$: 
\begin{equation}
\label{eq:eps_probs}
\tilde{f}_\sigma \propto f_\sigma q_\sigma^{\epsilon-1}
\end{equation}
for any $\eps $ with the normalization $\sum_\sigma \tilde{f}_\sigma =1$.
The  MLE solution for the re-weighted distribution is then $\tilde{p}^*_\sigma = \tilde{f}_\sigma$ following Eq.~(\ref{eq:MLEsolve}). 
Here, if we set $q_\sigma = p^*_\sigma$, then we have $\tilde{p}^*_\sigma \propto f^\epsilon_\sigma$, which becomes exactly flat for $\epsilon =0$.
In other words, complete erasure of the information in observed configurations implies knowing the true distribution $p^*_\sigma$.
Now the re-weighted model probability is
\begin{equation}
\label{eq:eps_model_probs}
\tilde{p}_\sigma \propto p_\sigma q_\sigma^{\epsilon-1} = p_\sigma^{\epsilon}
\end{equation}
with a specific choice of $q_\sigma = p_\sigma$.
Therefore, we obtain
\begin{equation}
\label{eq:eps}
\tilde p_\sigma = {\exp(- \eps E_\sigma) \over \tilde{Z}} \ \hbox{\rm with}\ \tilde{Z} \equiv \sum_\sigma  \exp(-\eps E_\sigma),
\end{equation}
where energy $E_\sigma = -w^IO_I(\sigma)$ or more specifically $E_\sigma = \sum_i h^i \sigma_i + \sum_{j<k} J^{jk} \sigma_j \sigma_k$ for the Ising model.
Here it is tempting to interpret $\epsilon$ as the usual inverse temperature $\beta$ in statistical mechanics. 
Although the final formula looks the same, the procedure to obtain the modified distribution is different: $\tilde{p}_\sigma \propto p_\sigma p_\sigma^{\epsilon -1} = p_\sigma^\epsilon$ versus $\tilde{p}_\sigma \propto (p_\sigma)^\beta = p_\sigma^\beta$.

Given the re-weighted data, we now want to find $\tilde{p}^*_\sigma$ that maximizes the likelihood ${\tilde{\cal L}} \equiv \prod_\sigma \tilde{p}^{\tilde{f_\sigma}}_\sigma$. 
The gradient of the logarithm of $\tilde{\cal L}$ is  
\begin{equation}
\label{eq:epsMgradient}
{\partial{\ln{\tilde{\cal L}}} \over {\partial w^{I}}} = \epsilon \langle O_I\rangle_{\tilde{f}} - \epsilon \langle O_I\rangle_{\tilde{p}}.
\end{equation}
Note that we consider the $w$ dependence only in $\tilde{p}_\sigma,$ and not in $\tilde{f}_\sigma,$ because we fix the re-weighting of the observed configurations. 
This modified gradient looks similar to the original gradient in Eq.~(\ref{eq:BMgradient}), but the situation is dramatically changed.
Now the two expectations in Eq.~(\ref{eq:epsMgradient}) are easily computable.
The first expectation still needs to consider only observed configurations $\hat{\sigma}$: 
\begin{equation}
\langle O_I \rangle_{\tilde{f}} = \frac{\sum_{\hat{\sigma}} O_I(\hat{\sigma}) \tilde{f}_{\hat{\sigma}}}{\sum_{\hat{\sigma}} \tilde{f}_{\hat{\sigma}}},
\end{equation}
because $\tilde{f}_\sigma \propto f_\sigma p_\sigma^{\epsilon-1} = 0$ for unobserved configurations $\sigma$. 
In the presence of $\epsilon$, the second expectation can be defined as
\begin{equation}
\langle O_I \rangle_{\tilde{p}} = \sum_{\sigma} O_I(\sigma) \tilde{p}_{\sigma} = \frac{\sum_\sigma O_I(\sigma) \exp(-\epsilon E_\sigma)}{\tilde{Z}}
\end{equation}
with $E_\sigma = -w^I O_I(\sigma)$ and $\tilde{Z}=\sum_\sigma \exp(\epsilon w^I O_I(\sigma))$.
Note $\partial \ln \tilde{Z}/ \partial w^I = \epsilon \langle O_I \rangle_{\tilde{p}}$.
Here the re-weighted partition function $\tilde{Z}$ can be expanded as follows to expose $\epsilon$ dependence:
\begin{widetext}
\begin{eqnarray}
\label{eq:Z}
\tilde{Z} &=& \sum_\sigma \exp(-\epsilon E_\sigma) = \sum_\sigma \exp(\sum_i \epsilon h^i \sigma_i + \sum_{j<k} \epsilon J^{jk} \sigma_j \sigma_k)  \\
&=& \sum_\sigma \prod_i \cosh(\epsilon h^i) \bigg[1+\sigma_i \tanh(\epsilon h^i) \bigg]  \prod_{j<k} \cosh(\epsilon J^{jk}) \bigg[1+\sigma_j \sigma_k \tanh(\epsilon J^{jk}) \bigg] \nonumber \\
&=& 2^M \prod_i \cosh(\epsilon h^i) \prod_{j<k} \cosh(\epsilon J^{jk}) \bigg[1+ \sum_{l<m} \tanh(\epsilon h^l) \tanh(\epsilon h^m) \tanh(\epsilon h^n)  \nonumber \\
&& \phantom{MMMMMMMMMMMMM}+ \sum_{l<m<n} \tanh(\epsilon J^{lm}) \tanh(\epsilon J^{ln}) \tanh(\epsilon J^{mn}) + \mathcal{O}(\epsilon^4) \bigg] \nonumber
\end{eqnarray}
\end{widetext}
Then, we truncate the logarithm of $\tilde{Z}$ up to order $\epsilon^2:$
\begin{equation}
\label{eq:lnZ}
\ln \tilde{Z} = M \ln 2 + \sum_i \ln \cosh(\epsilon h^i) + \sum_{j<k} \ln \cosh(\epsilon J^{jk}).  
\end{equation}
Finally, we obtain
${\partial \ln \tilde{Z}}/{\partial w^I} =  \epsilon \tanh(\epsilon w^I) = \epsilon^2 w^I$.
This leads to 
\begin{equation}
\langle O_I \rangle_{\tilde{p}} = \epsilon w^I.
\end{equation}
Thus, in the small $\epsilon$ limit, the expectation $\langle O_I \rangle_{\tilde{p}},$ which usually requires expensive calculations to deal with every configuration, does not require any calculation at all because it is simply $\epsilon w^I$.
This motivates us to call this learning algorithm the {\it erasure machine} or $\epsilon$-{\it machine}.
Finally, we derive an update algorithm for the model parameter $w_I$:
\begin{equation}
\label{eq:update}
\delta w^I =  \alpha \Big( \langle O_I \rangle_{\tilde{f}} - \epsilon w^I \Big)
\end{equation}
with an arbitrary learning rate $\alpha$.
This update rule is interesting from two perspectives.
First, the second term $-\epsilon w^I$ works as a regularizer for constraining the amplitude of $w^I$, which contributes to make this algorithm stable. This term may look similar to ridge regression but it is conceptually and computationally completely different as we introduced no regularization. Second, $\epsilon = 1$ leads to the Hopfield solution of $w^I = \langle O_I \rangle_f$ at the maximum likelihood condition ($\delta w^I = 0$).

Summarizing the learning algorithm of the $\epsilon$-machine,
\begin{enumerate}
\item[(i)] compute $p_\sigma (w)\propto \exp(w^I O_I(\sigma))$ initially with random $w$;
\item[(ii)]  re-weight $\tilde{f}_\sigma = f_\sigma p_\sigma^{\epsilon-1} / \sum_{\sigma'} f_{\sigma'} p_{\sigma'}^{\epsilon-1}$;
\item[(iii)] obtain $\langle O_I \rangle_{\tilde{f}} = \sum_{\hat{\sigma}} O_I(\hat{\sigma}) \tilde{f}_{\hat{\sigma}}$;
\item[(iv)] update $w^I \rightarrow w^I + \alpha \Big( \langle O_I \rangle_{\tilde{f}} - \epsilon w^I \Big)$;
\item[(v)] iterate (i)-(iv) 
\end{enumerate}

\begin{figure}
\centering
\includegraphics[width=9cm]{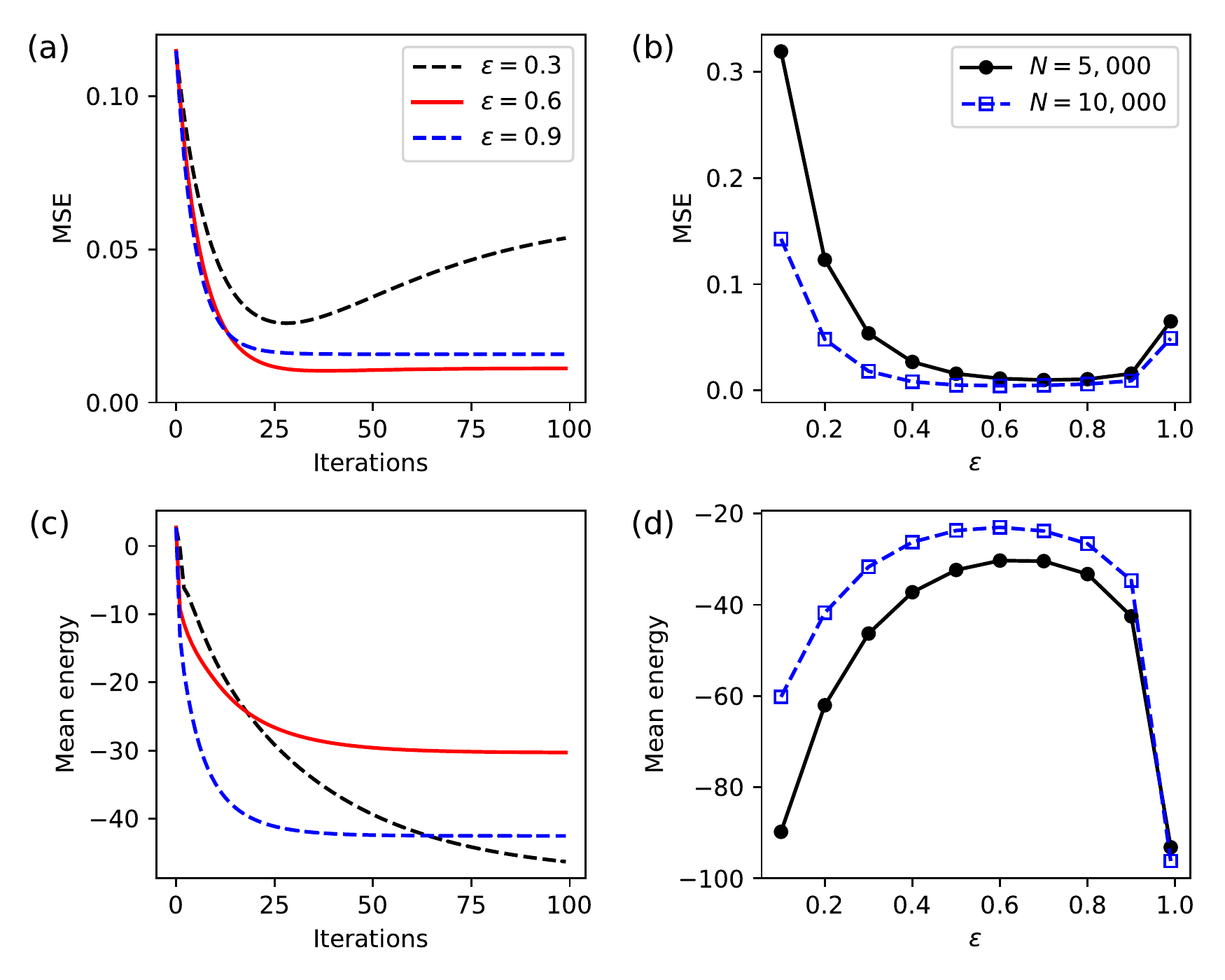}
\caption{ \label{fig:fig1} (Color online) Learning of the $\epsilon$-machine. (a) Mean squared error (MSE) between actual and inferred interactions and (c) mean energy of observed configurations for different $\epsilon$ values with $N=5,000$ during iterations. 
The minimal values of (b) MSE and (d) mean energy during iterations depending on $\epsilon$ values.
For the learning, we used a system with $M=40$ spins, and two sample sample sizes with $N = 5,000$ (filled black circles) and $N=10,000$ (empty blue squares).
}
\end{figure} 

{\it Results.} 
Now we demonstrate that the $\epsilon$-machine can efficiently infer model parameters for maximally explaining the distribution of observed configurations, especially in the regime where standard MLE is intractable.
For clear demonstration, we adopt an energy function of the Ising model, $E_\sigma = \sum_i h^i \sigma_i + \sum_{j<k} J^{jk} \sigma_j \sigma_k$, which has $L=M + M(M-1)/2$ parameters $w=\{h^i, J^{jk}\}$.
First, for simulating data, we randomly set the parameter values from a Gaussian distribution with zero mean and some variance that specifies bias and coupling strength, and define them as $w_{\text{true}}$.
We then generate observations $\hat{\sigma}$ from the distribution $p_\sigma(w_{\text{true}})$.
The goal of the inverse problem is to find values of these parameters, i.e., to recover $w_{\text{true}},$ in order to make the model distribution $p_\sigma = \exp(-E_\sigma)/Z$ close to the observed distribution $f_\sigma$ by examining only data $\{\hat{\sigma}\}.$

The $\epsilon$-machine can iteratively determine $w^I$ by using Eq.~(\ref{eq:update}).
As iteration goes on, inferred $w$ gets closer to $w_{\text{true}}$ as quantified by the mean squared error, MSE = $L^{-1} \sum_I (w^I - w^I_{\text{true}})^2$ (Fig.~\ref{fig:fig1}a).
For small $\epsilon$, too many iterations sometimes lead to worse inference. However, for large $\epsilon$, inference accuracy improves  during the iteration, and finally becomes saturated (Fig.~\ref{fig:fig1}a). 
The final value of MSE after training depends on the value of $\epsilon$  (Fig.~\ref{fig:fig1}b). 
Too small $\epsilon$ does not sufficiently regularize for determining $w^I$ in Eq.~(\ref{eq:update}), whereas $\epsilon$ too large can result in a poor approximation of the small $\epsilon$ expansion ($|\epsilon w^I| < 1$) in Eq.~(\ref{eq:Z}).
Therefore, it is important to find the optimal $\epsilon$ for the smallest MSE. Note that the MSE is analytic in $\epsilon$ when it is large enough to avoid $w$ values diverging, as mentioned above.  The flattening at the minimum of the MSE (Fig.~\ref{fig:fig1}b) then implies that the MSE is actually a constant in a range of $\epsilon.$
However, the MSE is not available for real data as $w_{\text{true}}$ is unknown, so we examined an alternative measure, the mean energy of observed configurations, $\langle E \rangle_f = \sum_{\hat{\sigma}} E_{\hat{\sigma}} f_{\hat{\sigma}}$, {\it without} re-weighting, which is also clearly analytic in $\epsilon$ for large enough $\epsilon.$
This mean energy decreased during iterations (Fig.~\ref{fig:fig1}c). 
While the final value of mean energy also depends on the value of $\epsilon$ (Fig.~\ref{fig:fig1}d),
for a range of  $\epsilon$, just as MSE became a constant, $\langle E \rangle_f$ shows minimal changes:
MSE becomes minimal, whereas $\langle E \rangle_f$ becomes maximal.  Initially the $\epsilon$-machine regards the observed sequences as very likely but as it learns the correct couplings, the unobserved sequences also factor into its updates, and the final average model probability of the observations decreases, or equivalently $\langle E \rangle_f$ becomes maximal.
Therefore, the $\epsilon$-machine first works in parallel and independently for a range of $\epsilon,$ then picks the optimal $\epsilon$ that maximizes $\langle E \rangle_f$. 
The stable range become larger as the number of samples increases (Fig.~\ref{fig:fig1}b and d).

\begin{figure}
\centering
\includegraphics[width=9cm]{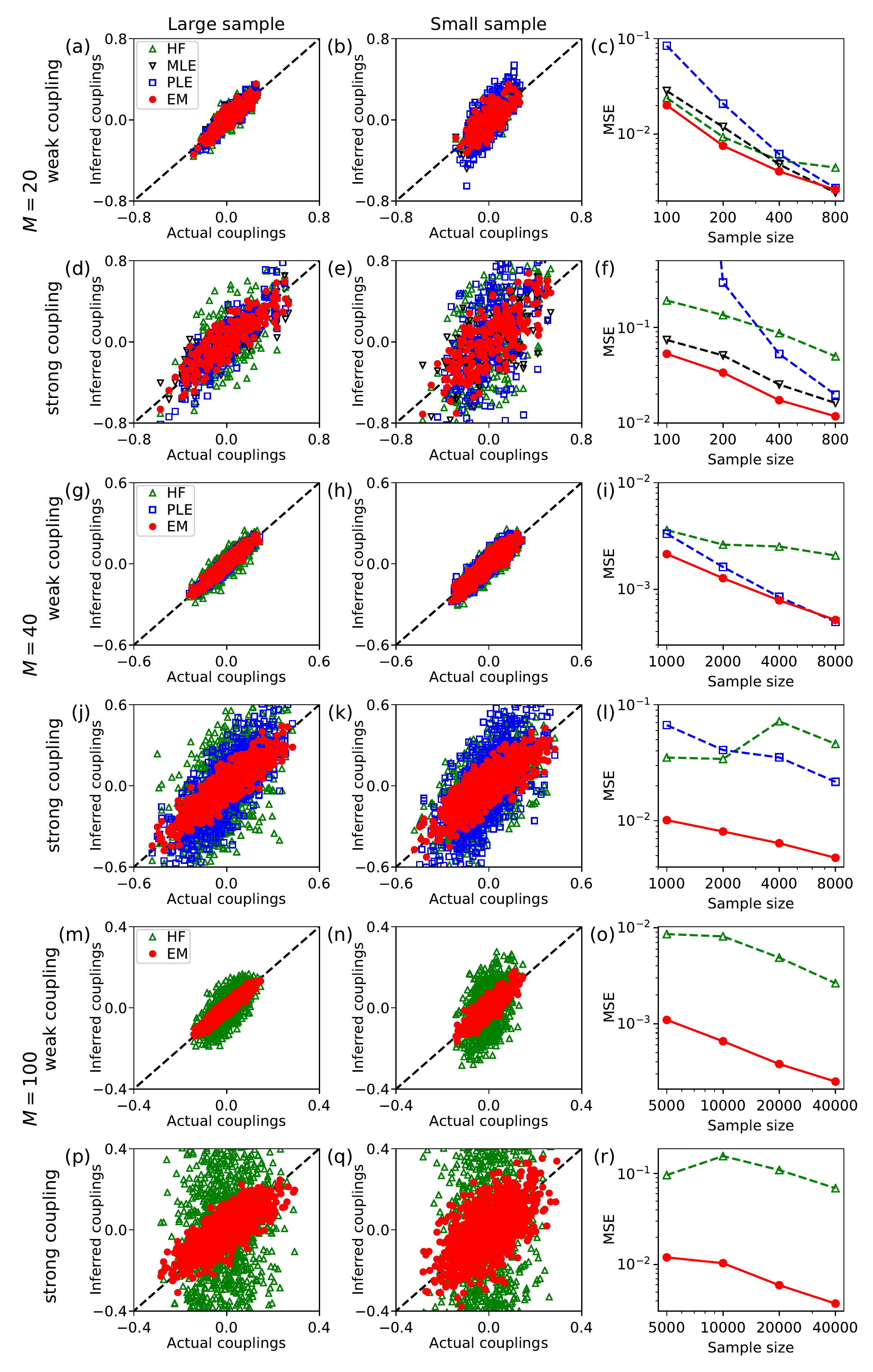}
\caption{ \label{fig:fig2} (Color online) Inference performance of the $\epsilon$-machine. Inferred interactions versus actual interactions for large (first column) and small sample sizes (second column), and for weak (odd row) and strong interactions (even row). Mean squared errors (MSE) between inferred  and actual interactions are plotted as a function of sample sizes (third column). Inference performances of Hopfield solution (HF), maximum likelihood estimation (MLE), pseudo-likelihood estimation (PLE), and $\epsilon$-machine (EM) are compared for various system sizes ($M=20, 40, 100$). Here MLE is not available for a larger systems ($M=40, 100$), and PLE is not available for the largest system ($M=100$).
}
\end{figure}

\begin{figure}
\centering
\includegraphics[width=5cm]{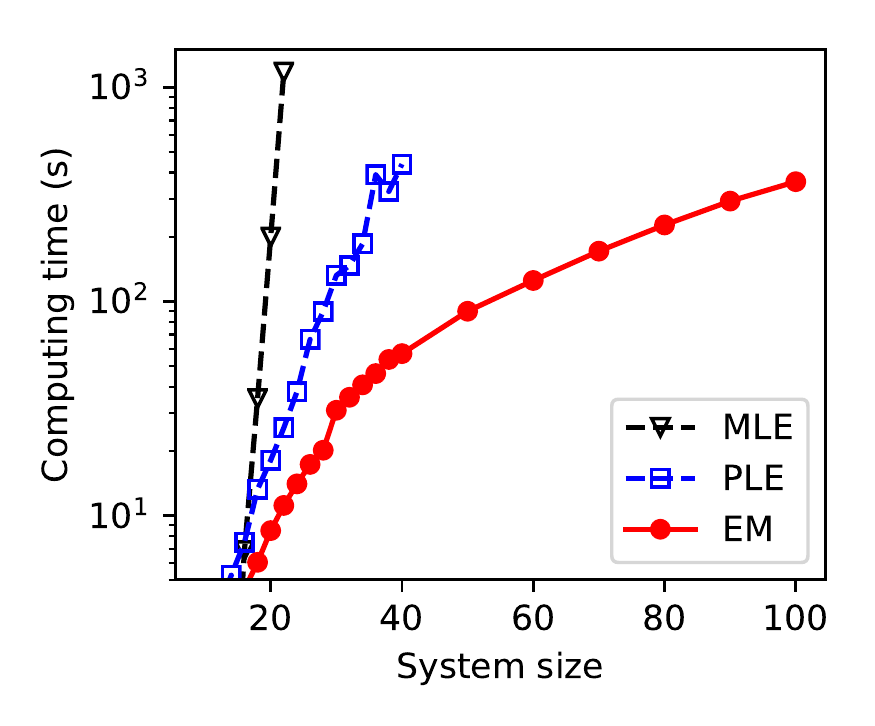}
\caption{ \label{fig:fig3} (Color online) Computing time of the $\epsilon$-machine. Computing time (seconds) is compared between $\epsilon$-machine (EM, filled red circles), maximum likelihood estimation (MLE, empty black triangles), and pseudo-likelihood estimation (PLE, empty blue squares) for different systems sizes given $N=10,000$ samples.
}
\end{figure}

We now compare the performance of the $\epsilon$-machine (EM) with existing methods (Fig.~\ref{fig:fig2}). For a small system ($M=20$), MLE can be used because the number of every possible configuration ($2^{10} \approx 10^6$) is computable. In the regime of weak interactions and large sample size, the EM performs as well as MLE and pseudo-likelihood estimation (PLE)~\cite{lee2019} (Fig.~\ref{fig:fig2}a).
On the other hand, the Hopfield solution (HF) is less accurate than the other three methods (Fig.~\ref{fig:fig2}c).
However, in the regime of strong interactions and/or small sample size, EM outperforms HF, MLE and PLE (Fig.~\ref{fig:fig2}f).
For a larger system ($M=40$), MLE is intractable because the partition function includes $2^{40}$ configurations. EM works as well as PLE in the limit of week interaction and large sample (Fig.~\ref{fig:fig2}i). However, EM works significantly better than PLE in the limit of strong interactions and/or small sample size (Fig.~\ref{fig:fig2}l). Nonetheless, HF does not do well compared with EM and PLE.
As the system size becomes much larger ($M=100$), only EM and HF can be used because PLE becomes intractable. EM still works well in this limit (Fig.~\ref{fig:fig2}m-r).
Furthermore, because EM only considers observed configurations, it significantly reduces computing time, compared with MLE and PLE. For $M = 20,$ EM takes approximately 25 times and 2 times less time than MLE and PLE, respectively. For $M=40$, EM takes approximately 8 times less time than PLE (Fig.~\ref{fig:fig3}).

\begin{figure}
\centering
\includegraphics[width=8cm]{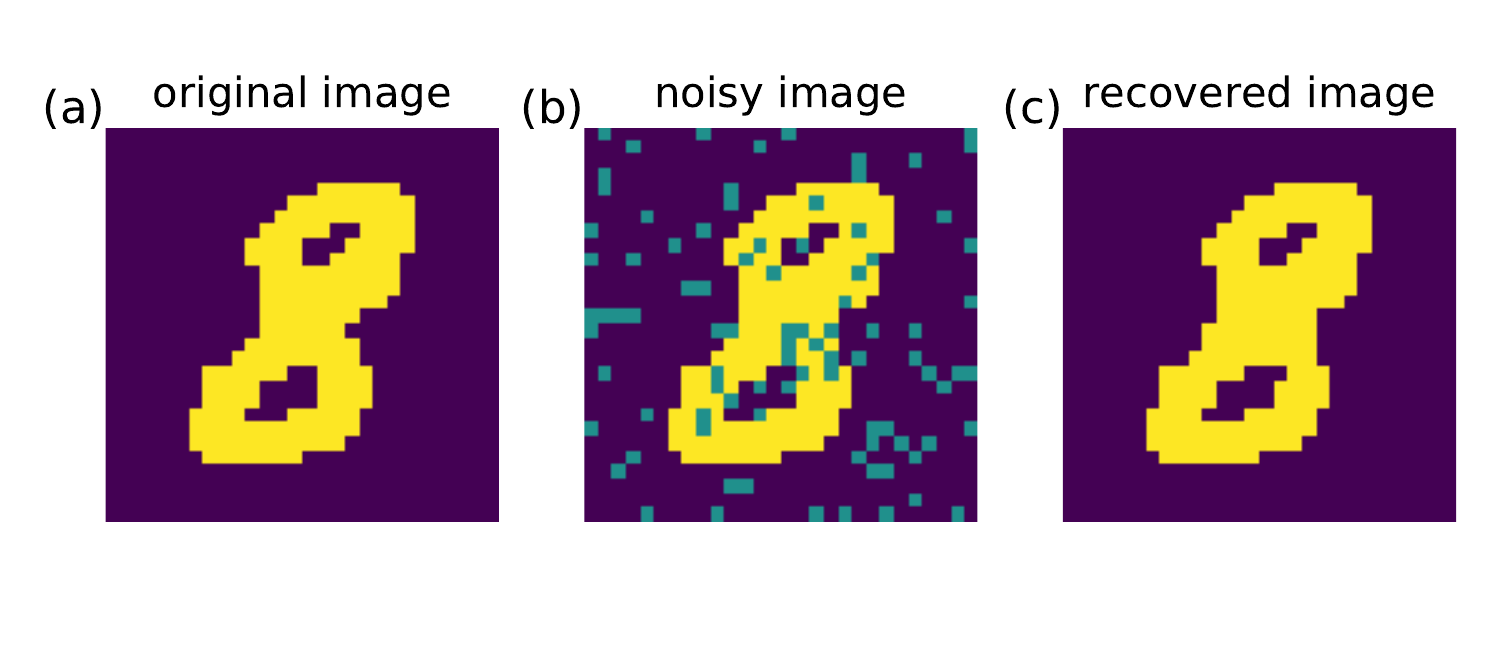}
\caption{ \label{fig:fig4} (Color online) Image reconstruction by the $\epsilon$-machine.
(a) An MNIST image. (b) 90 pixel values are missed (green pixels). (c) Recovered image with missing pixels reconstructed by the $\epsilon$-machine.
}
\end{figure}

Since the $\epsilon$-machine works effectively for large systems, we apply it to reconstruct missing pixels in real images.
We use the MNIST images of handwritten digits~\cite{Lecun1998}. 
The gray-scale values $x_i$ of $28 \times 28$ pixel images are binarized, $\sigma_{i} = 1$ for $x_i > 1$ or $\sigma_{i} =-1$ otherwise (Fig.~\ref{fig:fig4}a). Given a test image, we randomly select 90 pixels ($>$10 \% of total 784 pixels), and define them ($\sigma_i =0$) as missing pixels (Fig.~\ref{fig:fig4}b). 
Our goal is to reconstruct the missing pixels, and recover the original image.
Specifically, we use $N=5851$ samples of digit 8 in the MNIST training data. 
First, if the $i$-th pixel has a common value of $\sigma_i$ for more than 80\% of the training samples, the $i$-th missing pixel in the test image is simply reconstructed by the common $\sigma_i$ value.
However, the remaining $M=222$ pixels have a large sample variation. Therefore, we apply the $\epsilon$-machine to obtain $p_\sigma$ by inferring pixel bias and interactions.
Here we divide the pixel vector $\sigma = (\sigma_m, \sigma_m^c)$ into missing pixels $\sigma_m$ and observed pixels $\sigma_m^c$.
Then, by maximizing $p(\sigma_m|\sigma_m^c) \propto p(\sigma_m, \sigma_m^c) \equiv p_\sigma$, we can reconstruct $\sigma_m$ (Fig.~\ref{fig:fig4}c). 

{\it Discussion.} 
Inferring underlying models from observed configurations has become a cynosure with the present flood of big data.
However, big data is not yet big enough to use most available inference methods for large systems.
In this study, we proposed a data-driven algorithm for solving the inverse Ising problem without any assumption on the connectivity (e.g., weak coupling, sparse networks, no cycles, etc.) of underlying systems.
Unlike standard maximum likelihood estimation, our algorithm relies entirely on observed configurations with no need to sum over the vast number of unseen configurations. 
We systematically re-weighted the frequency of observed configurations to increase the entropy of the re-weighted observed configuration distribution, and in the process trivialized the computation of the exact partition function for every parameter value.
Since the $\epsilon$-machine requires only the  computation of  expectation values of observables in the re-weighted observed ensemble, it is very fast. Furthermore, it gives more accurate inference results than  state-of-the-art pseudo-likelihood methods  in the difficult inference regime of  limited sample size or strong coupling.

The concept of flattening the observed distribution and trivializing the partition function can be further extended to consider hidden variables~\cite{hoang2019b,roudi2011}, continuous variables~\cite{sohl2011, donner2017}, non-equilibrium asymmetric couplings~\cite{hoang2019a}, and  other inference problems where the computation of the partition function is unfeasible.

We thank Juyong Song for discussions during the initial stages of this study.
This work was supported by Intramural Research Program of the National Institutes of Health, NIDDK (D.-T.H., V.P.), and by the New Faculty Startup Fund from Seoul National University and the National Research Foundation of Korea (NRF) grant funded by the Korea government (MSIT) (No. 2019R1F1A1052916) (J.J.).

J.J. and D.-T.H. contributed equally to this work.

\bibliographystyle{apsrev}
\bibliography{e_machine}

\end{document}